\documentclass{article}

\usepackage{arxiv}

\usepackage[utf8]{inputenc} 
\usepackage[T1]{fontenc}    
\usepackage{hyperref}       
\usepackage{url}            
\usepackage{booktabs}       
\usepackage{amsfonts}       
\usepackage{nicefrac}       
\usepackage{microtype}      
\usepackage{lipsum}
\usepackage{multirow}
\usepackage{graphicx}

\title{Res-SE-Net: Boosting performance of Resnets by enhancing bridge-connections}

\author{
  Varshaneya V \\
  Department of Mathematics and Computer Science\\
  Sri Sathya Sai Institute of Higher Learning\\
  Prasanthi Nilayam, AP, India\\
  \texttt{varshaneya.v@gmail.com} \\
   \And
  Balasubramanian S \\
  Department of Mathematics and Computer Science\\
  Sri Sathya Sai Institute of Higher Learning\\
  Prasanthi Nilayam, AP, India\\
  \texttt{sbalasubramanian@sssihl.edu.in} \\
  \AND
  Darshan Gera \\
  Department of Mathematics and Computer Science\\
  Sri Sathya Sai Institute of Higher Learning\\
  Prasanthi Nilayam, AP, India\\
  \texttt{darshangera@sssihl.edu.in}
}

\begin{document}
\maketitle

\begin{abstract}
One of the ways to train deep neural networks effectively is to use residual connections. Residual connections can be classified as being either identity connections or bridge-connections with a reshaping convolution. Empirical observations on CIFAR-10 and CIFAR-100 datasets using a baseline Resnet model, with bridge-connections removed, have shown a significant reduction in accuracy. This reduction is due to lack of contribution, in the form of feature maps, by the bridge-connections. Hence bridge-connections are vital for Resnet. However, all feature maps in the bridge-connections are considered to be equally important. In this work, an upgraded architecture ``Res-SE-Net'' is proposed to further strengthen the contribution from the bridge-connections by quantifying the importance of each feature map and weighting them accordingly using Squeeze-and-Excitation (SE) block. It is demonstrated that Res-SE-Net generalizes much better than Resnet and SE-Resnet on the benchmark CIFAR-10 and CIFAR-100 datasets\footnote{Authors have uploaded their code \href{https://github.com/varshaneya/Res-SE-Net}{here}.}.
\end{abstract}

\keywords{Deep Residual learning \and Weighting activations \and Bridge-connections \and Resnet \and Squeeze-and-Excitation Net}

\section{Introduction}
Deep neural networks are increasingly being used in a number of computer vision tasks. One great disadvantage of training a very deep network is the ``vanishing gradient'' problem which delays the convergence. This is alleviated to some extent by initialization techniques mentioned in \cite{b12} and \cite{b14} and batch normalization \cite{b13}. It is observed by K. He et.al in \cite{b1} that, accuracy stagnates and degrades subsequently as the network becomes deeper. They argue that this degradation is not caused by over-fitting and that adding more layers to an already ``deep'' model results in an increase in train and test errors. Therefore in order to make training of deep networks possible, they introduce ``Resnets''. Resnet \cite{b1} attends to this problem by emphasizing on learning residual mapping rather than directly fit input to output. This is achieved by introducing skip connections which ensure a larger gradient is flown back during the back-propagation.

Subsequent to Resnet a plethora of variants like ResNeXt \cite{b19}, Densenet \cite{b20}, Resnet with stochastic depth \cite{b21} and preactivated Resnet \cite{b22} have been proposed that makes training very deep networks possible. All these variants have either focussed on pre-activation or split-transform-merge paradigm or dense skip connections or dropping layers at random. But none have investigated the bridge-connections in Resnet that connect two blocks with a varying number of feature maps. In this work, we investigate the effect of bridge-connections in Resnet and subsequently propose a network architecture called ``Res-SE-Net'' that performs better than the baseline Resnet and SE-Resnet.

\section{Related Work}
Training deep networks had been a concern until Resnets \cite{b1} were introduced. Resnet emphasize on learning residual mappings rather than directly fit input to output. Subsequent to Resnet, a lot of its variants have been proposed. Fully preactivated Resnet \cite{b22} performs activation before addition of identity to the residue to facilitate unhindered gradient flow through the shortcut connections to earlier layers. This makes training of a 1001 layered deep network possible. ResNeXt \cite{b19} follows a model which is like Inception net \cite{b23}, by splitting the input to Resnet block into multiple transformation paths and subsequently merging them before identity addition. The number of paths is a new hyperparameter that characterizes model capacity. With higher capacity, the authors have demonstrated improved performance without going much wider or deeper. Densenet \cite{b20} further exploits the effect of skip-connections by densely connecting through skip-connections the output of every earlier layer to every other following layer. The connections are made using depth concatenation and not addition. The authors argue that such connections would help in feature reuse and thereby an unhindered information flow. In \cite{b21}, the weight layers in the Resnet block are randomly dropped thereby only keeping skip-connections active in these layers. This gives rise to an ensemble of Resnets similar to dropout \cite{b24}. Dropping weight layers depend on ``survival probability''. This idea outperforms the baseline Resnet. Another important architecture that won the 2017 ILSVRC\footnote{http://image-net.org/challenges/LSVRC/} competition is SE-Resnet \cite{b2}. This winning architecture has in its base a Resnet with an SE block introduced between the layers of Resnet. This block quantifies the importance of feature maps instead of considering all of them equally likely. This has resulted in a significant level of improvement in performance of the Resnet.

Though Resnet has been studied in detail, to the best of our knowledge there has not been any work focusing on bridge-connections in Resnet. In this work, we investigate the effectiveness of bridge-connections and further propose a new architecture namely ``Res-SE-Net''. This architecture consists of an SE block in the bridge-connection to weight the importance of feature maps. Using the proposed architecture we demonstrate a superior performance on CIFAR-10 and CIFAR-100 benchmark datasets over baseline Resnet and SE-Resnet.

\section{Preliminaries}

\subsection{Resnet}

\begin{figure}[htbp]
\centerline{\includegraphics[scale=.4]{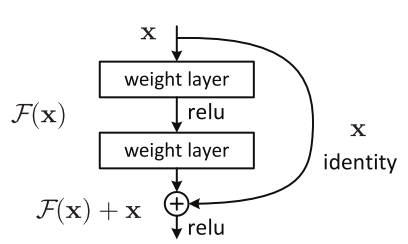}}
\caption{Resnet module (adapted from \cite{b1}).}
\label{basicResBlock}
\end{figure}

The idea behind Resnets \cite{b1} is to make a shallow architecture deeper by adding identity mapping from a previous layer to the current layer and then applying a suitable non-linear activation. Addition of skip-connections facilitates larger gradient flow to earlier layers thereby addressing the degradation problem as mentioned in \cite{b1}. The building block of a Resnet is depicted in Fig. \ref{basicResBlock}. Here x is identity and $\mathcal{F}($x$)$ is called the residual mapping.

\begin{figure}[htbp]
\centerline{\includegraphics[scale=.4]{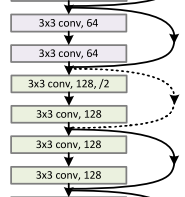}}
\caption{Bridge-connection in Resnet (adapted from \cite{b1}).}
\label{bridgeConnection}
\end{figure}

Resnet comprises of a stack of these blocks. A part of the 34-layer Resnet is shown in Fig. \ref{bridgeConnection}. The skip-connections that carry activations within a block are referred to as identity skip-connections and those that carry from block to another are called as bridge-connections. The dotted connection is an example of a bridge-connection. It involves a $1 \times 1$ convolution to increase the number of feature maps from 64 to 128 and also a downsampling operation to reduce their spatial dimension. In its absence, $\mathcal{F}($x$)$ and x will have incompatible dimensions to be added.

\subsection{Squeeze-and-Excitation Block}

Filters in a convolutional layer capture local spatial relationships in the form of feature maps. These feature maps are further used as they are without any importance being attached to them. In other words, each feature map is treated independent and equal. This may allow insignificant features that are not globally relevant to propagate through the network, thereby affecting the accuracy. Hence, to model the relationship between the feature maps, SE block is introduced in \cite{b2}. This enhances the quality of representations produced by a convolutional neural networks. SE block performs a recalibration of features so that the global information is used to weight features from the feature map that are more ``informative'' than the rest. 

\begin{figure}[htbp]
\centerline{\includegraphics[scale=.25]{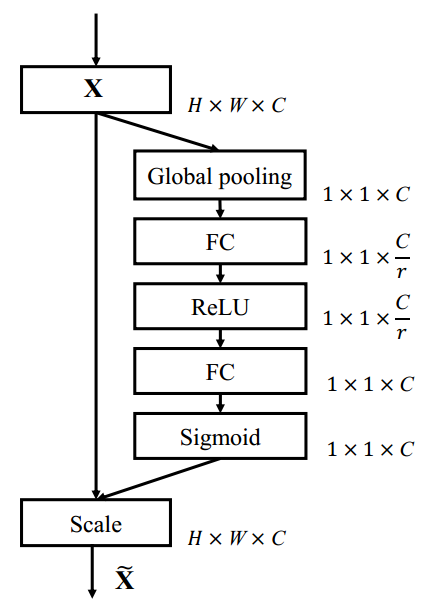}}
\caption{Squeeze-and-Excitation block (adapted from \cite{b2}).}
\label{SEblock}
\end{figure}

The SE block has two operations viz. squeeze and excitation. Features are first passed to the ``squeeze'' operation to produce a descriptor for each of the feature maps by aggregating along each of their spatial dimensions \cite{b2}. The descriptor produces an embedding of a global distribution of channel-wise feature responses. This allows information from the global receptive field of the network to be used by all its layers. The squeeze operation is followed by an ``excitation'' operation, wherein the embedding produced is used to get a collection of modulation weights for every feature map. These weights are applied to the feature maps to generate weighted feature maps as shown in Fig. \ref{SEblock}. In Fig. \ref{SEblock}, the input $X \in \mathbb{R}^{H \times W \times C}$ is fed to global pooling function which outputs a vector of dimension  $1 \times 1 \times C$. Its dimension is further reduced by $r$ using a fully-connected layer, which is followed by ReLU activation \cite{b17}. This constitutes a Squeeze operation. The output of squeeze operation is then upsampled to dimension $1 \times 1 \times C$ using another fully-connected layer followed by sigmoid activation which gives weights for each of the channels. This constitutes an Excitation operation. The input $X$ is thus rescaled using the output of excitation to get the weighted feature map which is $\tilde{X} \in \mathbb{R}^{H \times W \times C}$ as shown in Fig. \ref{SEblock}.

SE blocks add negligible extra computation and can be included in any part of the network \cite{b2}. SE-Resnet module is a Resnet module with each residual mapping passing through a SE block which is added with identity connection. SE-Resnet is a stack of SE-Resnet modules.

\section{Proposed Model} \label{modelSection}

Prior to the elucidation of the proposed model, we present the motivation. As shown in Fig. \ref{bridgeConnection}, bridge-connections (represented as dotted lines) connect two blocks of Resnet that have a different number of feature maps and different spatial dimensions. We now investigate the effectiveness of bridge-connections.

\subsection{Effect of Bridge-connections in Resnet}

Tables \ref{tabBaselineResnet} and \ref{tabNoBridgeResnet} compare the performance of various Resnet architectures with and without bridge-connections, respectively. The performance without bridge-connections drastically drops, particularly in Resnet-56 and Resnet-110. This comparison stresses the importance of bridge-connections.

\begin{table}[htbp]
\caption{Performance of baseline Resnets\cite{b1}}
\begin{center}
\begin{tabular}{|l|l|l|l|l|}
\hline
\multirow{2}{*}{Architecture} & \multicolumn{2}{l|}{CIFAR-10(Acc\%)} & \multicolumn{2}{l|}{CIFAR-100(Acc\%)} \\ \cline{2-5} 
 & Top-1 & Top-5 & Top-1 & Top-5 \\ \hline
Res-20 &91.4  &99.74  &67.37  &91.06  \\ \hline
Res-32 &92.32  &99.73  &69.8  &91.25  \\ \hline
Res-44 &93.57  &99.81  &73.15  &92.9  \\ \hline
Res-56 &93.16  &99.82  &73.8  &92.99  \\ \hline
Res-110 &93.66  &99.77  &73.33  &92.7  \\ \hline
\end{tabular}
\label{tabBaselineResnet}
\end{center}
\end{table}

However, in the original Resnet \cite{b1}, all feature maps in the bridge-connections are weighted equally. It is to be noted that SE-Resnet \cite{b2} weights the feature maps along the non-skip connections, based on their importance. The importance is learnt using a simple feed forward network that adds negligible computations. This idea motivated us to quantify the importance of feature maps that arise in bridge-connections.

\begin{table}[htbp]
\caption{Performance of baseline Resnets\cite{b1} without bridge-connections}
\begin{center}
\begin{tabular}{|l|l|l|l|l|}
\hline
\multirow{2}{*}{Architecture} & \multicolumn{2}{l|}{CIFAR-10(Acc\%)} & \multicolumn{2}{l|}{CIFAR-100(Acc\%)} \\ \cline{2-5} 
 & Top-1 & Top-5 & Top-1 & Top-5 \\ \hline
Res-20 &90.48  &99.71  &67.37  &91.06  \\ \hline
Res-32 &90.38  &99.62  &69.8  &91.25  \\ \hline
Res-44 &87.62  &99.51  &73.15  &92.9  \\ \hline
Res-56 &77.46  &98.55  &48.65  &76.67  \\ \hline
Res-110 &92.42  &99.83  &1.00  &5.00  \\ \hline
\end{tabular}
\label{tabNoBridgeResnet}
\end{center}
\end{table}

\subsection{Res-SE-Net - Our architecture}

\begin{figure}[]
\centerline{\includegraphics[scale=.35]{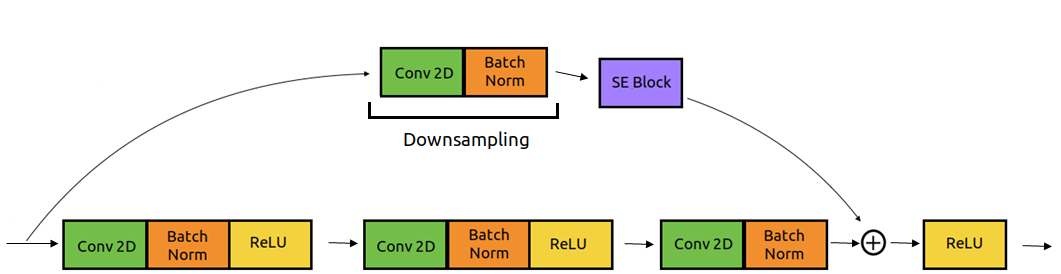}}
\caption{Modified bridge-connection \protect\footnotemark.}
\label{modifiedResblock}
\end{figure}

We incorporate an SE block in every bridge-connection in Resnet. Fig. \ref{modifiedResblock} shows an illustration of a modified bridge-connection. The proposed model Res-SE-Net has a similar architecture as mentioned in \cite{b1}. Specifically, our architecture is as follows.

The input image is of the size $32 \times 32$. The first layer is a $3 \times 3$ convolutional layer, which is followed by batch normalization \cite{b13} and ReLU \cite{b17}. This is followed by a stack of Resnet modules. A group of Resnet modules within the stack which have the same number of feature maps constitute a block. Average-pooling follows the stack of Resnet modules. The final layer is a fully-connected layer followed by softmax activation, which predicts the probability of an input belonging to a particular class. The sub-sampling of feature-maps is done in the first convolutional layer of every block, by performing the convolution with a stride of 2. There is a reduction in the size of feature maps and an increase in their number from one block to another. So to take activations from one block to another, the bridge-connection downsamples the feature map size and increases their number by using $1 \times 1$ convolutions with stride 2.

\footnotetext{Image from Andrew Ng's Deep Learning course and modified with an SE block.}

We add an SE block \cite{b2} on the bridge-connection just after downsampling. This ensures that when the feature maps are taken from one block to another they are weighted according to the content that they carry. Hence those features that are more relevant are given higher importance. The downsampled feature maps are sent from the previous block to the next one, so the weighting process must be done after downsampling in order to give more importance to the downsampled feature maps. Doing this before downsampling would reduce the significance of the weighted features. This is the primary reason for adding SE layer after downsampling and not before. Empirically we have also found that adding SE layer before the downsampling process gives less accuracy compared to adding it after downsampling.

\section{Experiments} \label{expt}

\subsection{Datasets}

We have used CIFAR-10 \cite{b18} and CIFAR-100 \cite{b18} datasets for all of our experiments. The CIFAR-10 dataset consists of 50000 training images and 10000 test images in 10 classes, with 5000 training images and 1000 test images per class. The CIFAR-100 dataset consists of 50000 training images and 10000 test images in 100 classes, with 500 training images and 100 test images per class. There are 20 main classes which contain these classes. The size of images in both the datasets is $32 \times 32$ and all of them are RGB images.

\subsection{Experimental setup}

We conducted our experiments on Resnets of depths 20, 32, 44, 56 and 110 layers. Our implementations are coded in Pytorch \cite{b16}. The code for baseline Resnet\footnote{Adapted from https://github.com/bearpaw/pytorch-classification.} and SE-Resnet\footnote{Adapted from https://github.com/moskomule/senet.pytorch.} have been adapted from existing implementations and modified. The following data augmentation techniques are used for training as mentioned in \cite{b1}:

\begin{itemize}
\item[$\bullet$] Padding with 4 pixels on each side.
\item[$\bullet$] Random cropping to a size of $32 \times 32$ from the padded image.
\item[$\bullet$] Random horizontal flip. 
\item[$\bullet$] Standard normalization. 
\end{itemize}

At the test time, we only normalize the images. The input to the network is of size $32 \times 32$.

The architecture of the network used, for both CIFAR-10 and CIFAR-100 datasets, is mentioned in section \ref{modelSection}. The training is started with an initial learning rate of 0.1 and subsequently, it is divided by 10 at 32000 and 48000 iterations. The training is done for a maximum of 64000 iterations. Stochastic Gradient Descent (SGD) is used for updating the weights. The weights in the model are initialized by the method described in \cite{b14} and batch normalization \cite{b13} is adopted. Dropouts \cite{b24} were not used for this model. The hyperparameters used are enlisted in Table \ref{hyperparameters}.

\begin{table}
\centering
\caption{Table of Hyperparameters}
\begin{tabular}{|l|l|} 
\hline
Hyperparameter        & Value   \\ 
\hline
Initial learning rate & 0.1     \\ 
\hline
Weight decay          & 0.0001  \\ 
\hline
Momentum              & 0.9     \\
\hline
Batch size            & 128     \\ 
\hline
\end{tabular}
\label{hyperparameters}
\end{table}

\section{Results}

Tables \ref{tabBaselineResnet}, \ref{tabBaselineSEResnet} and \ref{tabSEBridgeResnet} report the accuracies obtained by baseline Resnets, baseline SE-Resnets and Res-SE-Nets respectively. As evident from Table \ref{tabBaselineResnet}, the best performing Resnet is Resnet-110 with Top-1 accuracies of 93.66\% and 73.33\% on CIFAR-10 and CIFAR-100 respectively. Similarly, from Table \ref{tabBaselineSEResnet}, the best performing SE-Resnet is SE-Resnet-110 with Top-1 accuracies of 93.79\% and 72.99\% on CIFAR-10 and CIFAR-100 respectively. It is clear from Table \ref{tabSEBridgeResnet} that, our model, Res-SE-Net-110 reporting Top-1 accuracies of 94.53\% and 74.93\% on CIFAR-10 and CIFAR-100 datasets resspectively, significantly overwhelms the baseline Resnets and SE-Resnets. It can further be observed from Table \ref{tabSEBridgeResnet}, that SE-Resnet-44 performs exceedingly well compared to baseline Resnets and SE-Resnets. In fact, Res-SE-Net-44 is able to outperform Resnet-110 and SE-Resnet-110 by a significant margin of 0.42\% and 0.29\% on CIFAR-10 dataset respectively. On CIFAR-100 dataset, Res-SE-Net-44 dominates over Resnet-110 and SE-Resnet-110 by a margin of 0.5\% and 0.84\%. It is to be noted that Res-SE-Net-44 has 61.75\% and 62.06\% lesser number of parameters compared to Resnet-110 and SE-Resnet-110 respectively. Res-SE-Net-56 too exhibits outstanding performance on CIFAR-100 dataset compared to baseline Resnets and SE-Resnets, and near on-par performance on CIFAR-10 dataset with baseline Resnets and SE-Resnets. This strongly emphasizes the gravity of the proposed idea to activate the feature maps in bridge-connections by their importance. The proposed idea enables a reasonably deep network with lesser number of parameters to outperform very deep networks.

\begin{table}[htbp]
\caption{Performace of baseline SE-Resnets\cite{b2}}
\begin{center}
\begin{tabular}{|l|l|l|l|l|}
\hline
\multirow{2}{*}{Architecture} & \multicolumn{2}{l|}{CIFAR-10(Acc\%)} & \multicolumn{2}{l|}{CIFAR-100(Acc\%)} \\ \cline{2-5} 
 & Top-1 & Top-5 & Top-1 & Top-5 \\ \hline
SE-Resnet-20 &92.1  &99.7  &68.31  &90.69  \\ \hline
SE-Resnet-32 &93.08  &99.83  &70.09  &92.12  \\ \hline
SE-Resnet-44 &93.71  &99.77  &71.2  &91.78  \\ \hline
SE-Resnet-56 &93.64  &99.76  &71.42  &91.28  \\ \hline
SE-Resnet-110 &93.79  &99.74  &72.99  &92.71  \\ \hline
\end{tabular}
\label{tabBaselineSEResnet}
\end{center}
\end{table}

\begin{table}[htbp]
\caption{Performance of proposed model}
\begin{center}
\begin{tabular}{|l|l|l|l|l|}
\hline
\multirow{2}{*}{Architecture} & \multicolumn{2}{l|}{CIFAR-10(Acc\%)} & \multicolumn{2}{l|}{CIFAR-100(Acc\%)} \\ \cline{2-5} 
 & Top-1 & Top-5 & Top-1 & Top-5 \\ \hline
Res-SE-Net-20 &91.9  &99.75  &67.99  &91.03  \\ \hline
Res-SE-Net-32 &92.79  &99.82  &69.93  &91.95  \\ \hline
Res-SE-Net-44 &94.08  &99.83  &73.83  &93.54  \\ \hline
Res-SE-Net-56 &93.64  &99.87  &74.29  &93.45  \\ \hline
Res-SE-Net-110 &94.53  &99.87  &74.93  &93.48  \\ \hline
\end{tabular}
\label{tabSEBridgeResnet}
\end{center}
\end{table}

The improvement in accuracies of Res-SE-Nets, for both the datasets in comparison to baseline Resnets \cite{b1} and SE-Resnets \cite{b2} are tabulated in Table \ref{tabSEBridgeResnetImprovement} and Table \ref{tabSEBridgeSEResnetImprovement} respectively. Res-SE-Net outperforms baseline Resnet by 0.566\% i.e. about 56 images on CIFAR-10 and by 0.704\% i.e. about 70 images on CIFAR-100 datasets on an average. Res-SE-Net-110 has achieved the maximum improvement in accuracy of 0.87\% on CIFAR-10 and 1.6\% on CIFAR-100 over Res-110. Similarly, Res-SE-Net outperforms SE-Resnet by 0.124\% on CIFAR-10 and by 1.392\% on CIFAR-100 datasets respectively. Res-SE-Net-110 has achieved maximum overall improvement over SE-Resnet-110 with an increase in accuracy of 0.74\% on CIFAR-10 and of 1.94\% on CIFAR-100 datasets respectively.

\begin{table}[!htb]
\caption{Improvement in performance from baseline Resnet}
\begin{center}
\begin{tabular}{|l|l|l|l|l|}
\hline
\multirow{2}{*}{Architecture} & \multicolumn{2}{l|}{CIFAR-10(\%)} & \multicolumn{2}{l|}{CIFAR-100(\%)} \\ \cline{2-5} 
 & Top-1 & Top-5 & Top-1 & Top-5 \\ \hline
Res-SE-Net-20 &0.5  &0.01  &0.62  &-0.03  \\ \hline
Res-SE-Net-32 &0.47  &0.09  &0.13  &0.7  \\ \hline
Res-SE-Net-44 &0.51  &0.02  &0.68  &0.55  \\ \hline
Res-SE-Net-56 &0.48  &0.05  &0.49  &0.46  \\ \hline
\textbf{Res-SE-Net-110} &\textbf{0.87} &\textbf{0.1}  &\textbf{1.6}  &\textbf{0.78}  \\ \hline
Average Improvement &0.566 &0.504  &0.704  &0.492  \\ \hline
\end{tabular}
\label{tabSEBridgeResnetImprovement}
\end{center}
\end{table}

\begin{table}[!htb]
\caption{Improvement in performance from baseline SE-Resnet}
\begin{center}
\begin{tabular}{|l|l|l|l|l|}
\hline
\multirow{2}{*}{Architecture} & \multicolumn{2}{l|}{CIFAR-10(\%)} & \multicolumn{2}{l|}{CIFAR-100(\%)} \\ \cline{2-5} 
 & Top-1 & Top-5 & Top-1 & Top-5 \\ \hline
Res-SE-Net-20 &-0.2  &0.05  &-0.32  &0.34  \\ \hline
Res-SE-Net-32 &-0.29  &-0.01  &-0.16  &-0.17  \\ \hline
\textbf{Res-SE-Net-44} &\textbf{0.37}  &\textbf{0.06}  &\textbf{2.63}  &\textbf{1.76}  \\ \hline
Res-SE-Net-56 &0.0  &0.11  &2.87  &2.17  \\ \hline
\textbf{Res-SE-Net-110} &\textbf{0.74} &\textbf{0.13}  &\textbf{1.94}  &\textbf{0.77}  \\ \hline
Average Improvement &0.124 &0.068  &1.392  &0.974  \\ \hline
\end{tabular}
\label{tabSEBridgeSEResnetImprovement}
\end{center}
\end{table}

With the improvement that addition of SE block provides, one might want to add SE blocks to all of the skip-connections so as to make the performance even better. But we have empirically found that adding an SE block to every identity skip-connections degrades the performance on CIFAR-10 and CIFAR-100 datasets as the depth increases. Also, as for the reasons mentioned in section \ref{modelSection}, the addition of SE block before downsampling does not give better results either.

\begin{figure}[!htb]
\centerline{\includegraphics[scale=.3]{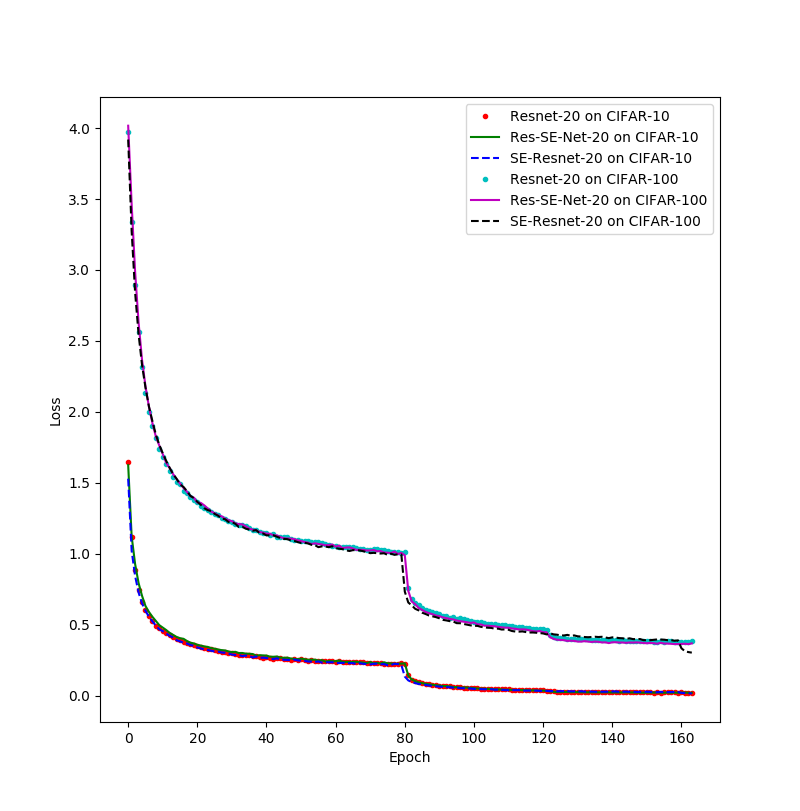}
\includegraphics[scale=.3]{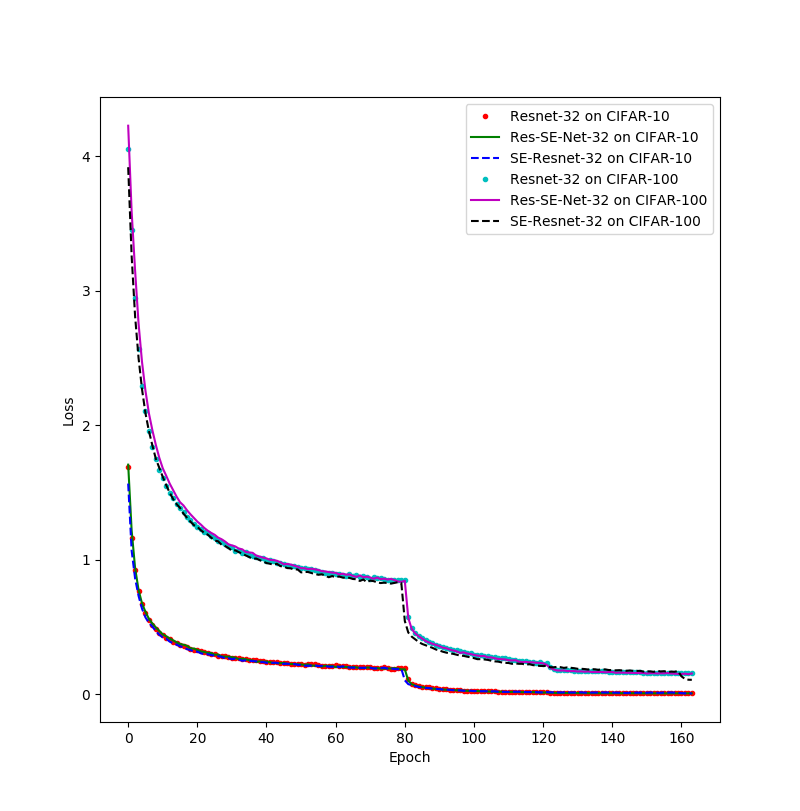}}
\caption{Left: Training losses plotted for depth of 20 layers. Right: Training losses plotted for depth of 32 layers.}
\label{res2032loss}
\end{figure}

\begin{figure}[!htb]
\centerline{\includegraphics[scale=.3]{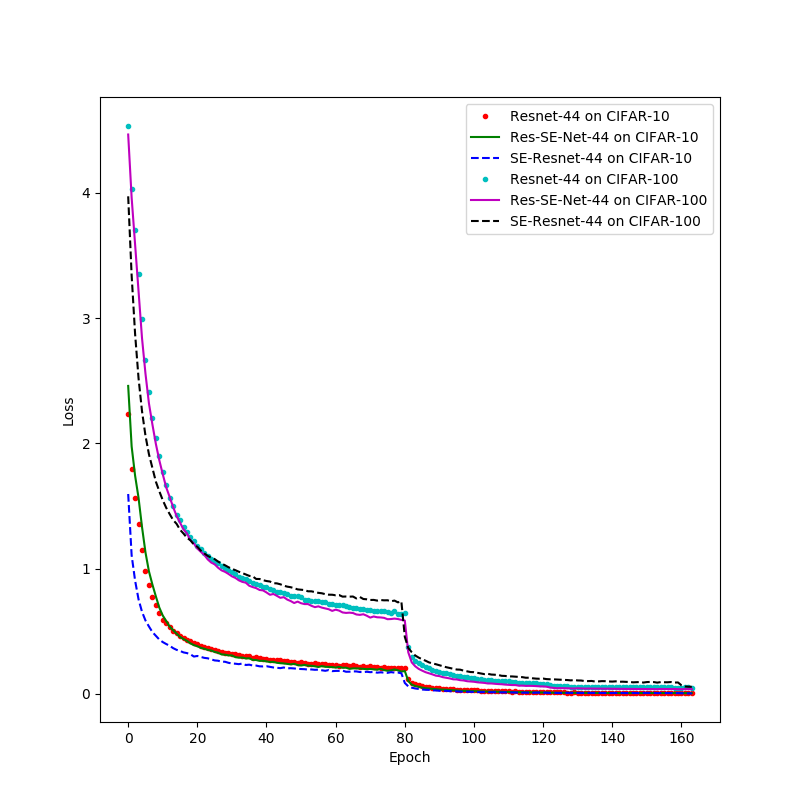}
\includegraphics[scale=.3]{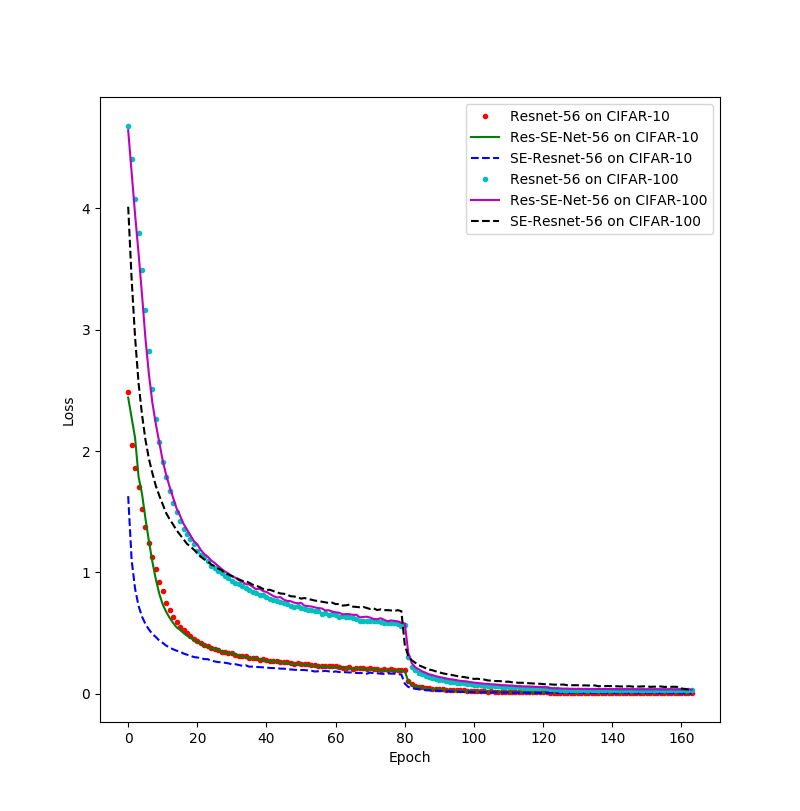}}
\caption{Left: Training losses plotted for depth of 44 layers. Right: Training losses plotted for depth of 56 layers.}
\label{res4456loss}
\end{figure}

\begin{figure}[!htb]
\centerline{\includegraphics[scale=.3]{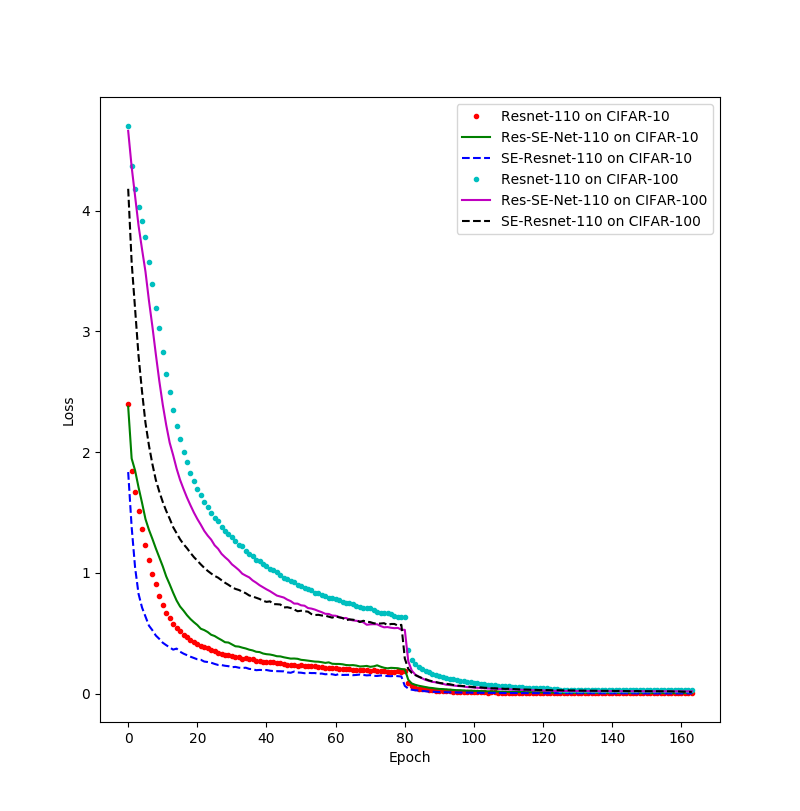}}
\caption{Training losses plotted for depth of 110 layers.}
\label{res110loss}
\end{figure}

We now analyze the training phase of Res-SE-Net by plotting training losses for all of the aforesaid depths and for both the datasets. From Fig. \ref{res2032loss}, Fig. \ref{res4456loss} and Fig. \ref{res110loss} we can conclude that training of Res-SE-Net has taken place smoothly. There is no abrupt increase in training loss of Res-SE-Net models. This shows that gradient flow has not been hindered by the introduction of an SE block in bridge-connections, keep intact the principle of Resnet (base of our Res-SE-Net) that skip-connections facilitate smooth training of deep networks.

\section{Conclusion}

In this work, we proposed a new architecture named ``Res-SE-Net'' that makes bridge-connections in Resnets more influential. This is achieved by incorporating an SE block in every bridge-connection. Res-SE-Net surpassed the performances of baseline Resnet and SE-Resnets by significant margins on CIFAR-10 and CIFAR-100 datasets. Further, we demonstrated that reasonably sized deep networks with positively contributing bridge-connections can outperform very deep networks. Also, we illustrated that addition of an SE block does not affect training. In future, we would like to explore other ways of making bridge-connections in Resnets influential towards enhancement in performance.

\section{Acknowledgment}
The authors wish to dedicate this work to the founder chancellor of Sri Sathya Sai Institute of Higher Learning, Bhagawan Sri Sathya Sai Baba. The authors also wish to extend their gratitude to Dr. Vineeth N Balasubramanian, Associate Professor in the Department of Computer Science and Engineering, Indian Institute of Technology - Hyderabad.

\bibliographystyle{unsrt}  


\end{document}